\documentclass[12pt]{article}
\pdfoutput=1
\usepackage[utf8]{inputenc}
\usepackage[T1]{fontenc}
\usepackage{float}
\usepackage{authblk}
\usepackage[normalem]{ulem}
\usepackage{xcolor}
\usepackage{amsmath}
\usepackage{amsthm}
\usepackage{amssymb}
\usepackage{adjustbox}
\usepackage{graphicx}
\usepackage{booktabs}
\usepackage{mathtools}
\usepackage{fullpage}
\usepackage[mathlines]{lineno}
\usepackage{multirow}
\usepackage{subcaption}
\usepackage[style=numeric-comp,sorting=none,maxnames=99]{biblatex}

\usepackage[colorlinks = true,
            linkcolor = black,
            urlcolor  = black,
            citecolor = black]{hyperref}

\title{Designing explainable artificial intelligence with active inference: A framework for transparent introspection and decision-making}

\author[1,2]{Mahault~Albarracin}
\author[1,3,4]{Inês~Hipólito}
\author[1,5]{Safae~Essafi Tremblay} 
\author[1]{Jason~G.~Fox}
\author[1]{Gabriel~René}
\author[1,6]{Karl~Friston}
\author[1,6]{Maxwell~J.~D.~Ramstead}

\affil[1]{VERSES Research Lab, Los Angeles, CA 90016, USA}
\affil[2]{Département d'informatique, Université du Québec à Montréal, 201, Avenue du Président-Kennedy, Montréal, H2X 3Y7}
\affil[3]{Berlin School of Mind \& Brain, Humboldt-Universität zu Berlin, Berlin, Germany}
\affil[4]{Department of Philosophy, Macquarie University, Sydney, New South Wales, Australia}
\affil[5]{Département de philosophie, Université du Québec à Montréal, 455, Boulevard René-Lévesque Est, Montréal, H2L 4Y2}
\affil[6]{Wellcome Centre for Human Neuroimaging, University College London,\protect\\London WC1N 3AR, UK}

\date{\today}

\begin{document}

\maketitle


\begin{abstract}

This paper investigates the prospect of developing human-interpretable, explainable artificial intelligence (AI) systems based on active inference and the free energy principle. We first provide a brief overview of active inference, and in particular, of how it applies to the modeling of decision-making, introspection, as well as the generation of overt and covert actions. We then discuss how active inference can be leveraged to design explainable AI systems, namely, by allowing us to model core features of ``introspective'' processes and by generating useful, human-interpretable models of the processes involved in decision-making. We propose an architecture for explainable AI systems using active inference. This architecture foregrounds the role of an explicit hierarchical generative model, the operation of which enables the AI system to track and explain the factors that contribute to its own decisions, and whose structure is designed to be interpretable and auditable by human users. We outline how this architecture can integrate diverse sources of information to make informed decisions in an auditable manner, mimicking or reproducing aspects of human-like consciousness and introspection. Finally, we discuss the implications of our findings for future research in AI, and the potential ethical considerations of developing AI systems with (the appearance of) introspective capabilities.

\end{abstract}


\tableofcontents 

\vskip0.5cm 

\subsection*{Acknowledgements}


The authors are grateful to VERSES for supporting the open access publication of this paper. SET is supported in part by funding from the Social Sciences and Humanities Research Council of Canada (Ref: 767-2020-2276). KF is supported by funding for the Wellcome Centre for Human Neuroimaging (Ref: 205103/Z/16/Z) and a Canada-UK Artificial Intelligence Initiative (Ref: ES/T01279X/1). The authors are grateful to Brennan Klein for assistance with typesetting. 

\subsection*{Conflict of interest statement}

The authors disclose that they are contributors to the Institute of Electrical and Electronics Engineers (IEEE) P2874 Spatial Web Working Group.

\pagebreak

\section{Introduction: Explainable AI and active inference}

Artificial intelligence (AI) systems continue to proliferate and, at the time of writing, have become an integral part of various intellectual and industrial domains, including healthcare, finance, and transportation \cite{raghupathi2014big, mascarenhas2023promise}. Traditional AI models, such as deep learning neural networks, have been widely recognized for their ability to achieve high performance and accuracy across various tasks \cite{goodfellow2016deep, lecun2015deep}. However, it is well known that these models almost invariably function as ``black boxes,'' with limited transparency and interpretability of their decision-making processes \cite{castelvecchi2016can, gunning2017explainable}. This lack of explainability can lead to skepticism and reluctance to adopt AI systems---and indeed, to harm, particularly in high-stakes situations, where the consequences of a wrong decision can be severe and harmful \cite{doshi2017towards, ribeiro2016should, birhane2021impossibility, birhane2022frameworks}. Indeed, a lack of explainability precludes applications in certain domains, such as fintech.

The problem of explainable AI (sometimes referred to as the ``black box'' problem) is the problem of understanding and interpreting how these models arrive at their decisions or predictions \cite{belisle2022artificial, bauer2021expl}. While researchers and users may have knowledge of the inputs provided to the model and the corresponding outputs that it produces, comprehending the internal workings and decision-making processes of AI systems can be complex and challenging. This is in no small part because their intricate architectures and numerous interconnected layers learn to make predictions by analyzing vast amounts of training data and adjusting their internal parameters, without explicit instruction from a programmer \cite{ali2023explainable}. The method by which these systems are trained thus, by design, limits their explainability. Moreover, the internal computations that are performed by these models---when they engage in decision-making---can be highly complex and nonlinear, making it difficult to extract meaningful explanations of their behavior, or insights into their decision-making process \cite{esterhuizen2022interpretable}. This problem is compounded by the fact that most machine learning implementations of AI fail to represent or quantify their uncertainty; especially, uncertainty about the parameters and weights that underwrite their accurate performance. This means that AI, in general, cannot evaluate (or report) the confidence in its decisions, choices or recommendations.

The lack of interpretability poses several challenges. Firstly, it hampers transparency and makes audits by third parties next to impossible, as the designers, users, and stakeholders of these systems may struggle to understand why a particular decision or prediction was made. This becomes problematic in critical domains such as healthcare or finance, where the ability to explain the reasoning behind a decision is essential for trust, accountability, and compliance with regulations \cite{mishra2021transparent, von2021transparency}. Secondly, the black box nature of machine learning models can hinder the identification and mitigation of biases or discriminatory patterns. Without visibility into the underlying decision-making process, it becomes challenging to detect and address biases that may exist within the model's training data or architecture. 

This opacity can lead to unfair or biased outcomes, perpetuating social inequalities or discriminatory practices \cite{veale2017fairer, van2022overcoming, nascimento2023comparing}. Additionally, the lack of interpretability of the model limits its ability to provide meaningful explanations to end-users. Individuals interacting with machine learning systems often seek explanations for the decisions made by these systems \cite{stiglic2020interpretability, laato2022explain}. For instance, in medical diagnosis, patients and healthcare professionals may want to understand why a particular diagnosis or treatment recommendation was given \cite{neri2023explainable, oberste2023designing}; or consider automated suggestions in practical industrial settings \cite{le2023exploring}. Without explainability, users may be hesitant to trust the system's recommendations or may feel apprehensive (not without good reason) about relying on the outputs of such models. 

Accordingly, the need for explainable AI has become increasingly important \cite{adadi2018peeking}. ``Explainable AI'' refers to the development of AI systems that can provide human-understandable explanations for their decisions and actions \cite{guidotti2018survey}. This level of transparency is crucial for fostering trust \cite{burrell2016machine}, ensuring accountability \cite{miguel2021putting}, and facilitating inclusive collaboration between humans and AI systems \cite{kokciyan2021sociotechnical, hipolito2023enactive, birhane2022forgotten}. Recent efforts to regulate AI may turn explainability into a requirement for the deployment of any AI system at scale. For instance, in the United States, the National Institute of Standards and Technology (NIST) released its Artificial Intelligence Risk Management Framework (RMF) in 2023, which includes explainability and interpretability as crucial characteristics of a trustworthy AI system. The RMF is envisioned as a guide for tech companies to manage the risks of AI and could eventually be adopted as an industry standard. In a similar vein, US Senator Chuck Schumer has led a congressional effort to establish US regulations on AI, with one of the key aspects being the availability of explanations for how AI arrives at its responses \cite{drake2023euai}.

In the European Union, a proposed Regulation Laying Down Harmonized Rules on Artificial Intelligence (better known as the ``AI Act'') is set to increase the transparency required for the use of so-called ``high-risk'' AI systems. For instance, groups that deploy automated emotion recognition systems may be obligated to inform those on whom the system is being deployed that they are being exposed to such a system. The AI Act is expected to be finalized and adopted in 2023, with its obligations likely to apply within three years’ time. The Council of Europe is also in the process of developing a draft convention on artificial intelligence, human rights, democracy, and the rule of law, which will be the first legally binding international instrument on AI. This convention seeks to ensure that research, development, and deployment of AI systems are consistent with the values and interests of the EU, and that they remain compatible with the AI Act and the proposed AI Liability Directive, which includes a risk-based approach to AI. In addition, the US-EU Trade and Technology Council published a joint Roadmap for Trustworthy AI and Risk Management in 2022, which aims to advance collaborative approaches in international standards bodies related to AI, among other objectives \cite{skeath2023ftc}. Therefore, explainability is clearly a major issue in research, development, and deployment of AI systems, and will remain so for the foreseeable future. 

Explainable AI aims to bridge the gap between the complexity and lack of auditability of contemporary AI systems and the need for human interpretability and auditability \cite{adadi2018peeking, guidotti2018survey, brennen2020people}. It seeks to provide insights into the factors that influence AI decision-making, enabling users to understand the explicit reasoning and other factors driving the output of AI systems. Understanding the performance and potential biases of AI systems is crucial for their ethical and responsible deployment \cite{ratti2022explainable, ridley2022explainable}. This understanding, however, must extend beyond the performance of AI systems on academic benchmarks and tasks to include a deep understanding of what the models represent or learn, as well as the algorithms that they instantiate \cite{guest2023logical}.

Transparency considerations are embedded in the design, development, and deployment of AI systems, from the societal problems that arise worth developing a solution, to the data collection stage, and still at the point where the AI system is deployed in the real world and iteratively improved \cite{hipolito2023enactive, hipolito2023human}. This transparency may enable the implementation of other ethical AI dimensions like interpretability, accountability, and safety \cite{chaudhry2022transparency}. 

Researchers have been exploring various approaches to develop more explainable AI systems \cite{arrieta2020explainable, doshi2017towards}. However, these efforts have yet to yield a principled and widely accepted path method for, or path to, explainability. One promising direction is to draw inspiration from research into human introspection and decision-making processes. Furthermore, a two-stage decision-making process, which includes a reflection stage where the network reflects on its feed-forward decision, can enhance the robustness and calibration of AI systems \cite{prabhushankar2022introspective}. It has been suggested that explainability in AI systems can be further enhanced through techniques such as layer-wise relevance propagation \cite{bach2015pixel} and saliency maps \cite{zhang2018interpretable}, which aid in visualizing the model's reasoning process. By translating the internal models of AI systems into human-understandable explanations, we can foster trust and collaboration between AI systems and their human users \cite{lamberti2023overview}. However, as \cite{guest2023logical} argue, we must also consider the metatheoretical calculus that underpins our understanding and use of these models. This involves not only considering the performance of the model on a task, but also the implications of the performance of the model for our understanding of the mind and brain.

In this paper, we investigate the potential of active inference, and the free energy principle (FEP) upon which is based \cite{ramstead2022bayesian, friston2022designing}, to enhance explainability in AI systems, notably by capturing core aspects of introspective processes, hierarchical decision-making processes, and (cover and overt) forms of action in human beings \cite{hohwy2013predictive, ramstead2023inner, ramstead2023mum}. The FEP is a variational principle of information physics that can be used to model the dynamics of self-organizing systems like the brain. Active inference is an application of the FEP to model the perception-action loops of cognitive systems: it provides us with the basis of a unified theory of the structure and function of the brain (and indeed, of living and self-organizing systems more generally; \cite{ramstead2018answering, ramstead2019variational}. Active inference allows us to model self-organizing systems like brains as being driven by the imperative to minimize surprising encounters with the environment; where this surprise scores how far a thing or system deviates from its characteristic states (e.g., a fish out of water). By doing so, the brain continually updates and refines its world model, allowing the agent to act adaptively and in situationally appropriate ways.

The relevance of using active inference is that the models of cognitive dynamics---and in particular, introspection---that have been developed using its tools can be adapted to enable the design of human interpretable and auditable (and indeed, self-auditable) AI systems. The ethical and epistemological or epistemic gains that this enables are notable. The proposed active inference based AI system architecture would enable artificial agents to access and analyze their own internal states and decision-making processes, leading to a better understanding of their decision-making processes, and the ability to report on themselves. Proof of concept for this kind of ``self report'' is already at hand \cite{parr2021understanding} and, in principle, is supported in any application of active inference. At one level, committing to a generative model---implicit in any active inference scheme---dissolves the explainability problem. This is because one has direct access to the beliefs and belief-updating of the agent in question. 

Indeed, this is why active inference has been so useful in neuroscience to model and explain behavioral and neuronal responses in terms of underlying belief states: e.g., \cite{smith2021active, smith2021computational, adams2013predictions, adams2022everything, sterzer2018predictive}. As demonstrated in \cite{parr2021understanding} it is a relatively straightforward matter to augment generative models to self-report their belief states. In this paper, we address a slightly more subtle aspect of explainability that rests upon ``self-access''; namely, when an agent infers its own ``states of mind''---states of mind that underwrite its sense-making and choices. Crucially, this kind of meta-inference \cite{fleming2020awareness, yon2021precision, frith2023consciousness, sandved2021towards} may rest on exactly the representations of uncertainty (a.k.a., precision) that are absent in conventional AI.

This paper is organized as follows. We first introduce essential aspects of active inference. We then discuss how active inference can be used to design explainable AI systems. In particular, we propose that active inference can be used as the basis for a novel AI architecture---based on explicit generative models---that both endows AI systems with a greater degree of explainability and audibility from the perspective of users and stakeholders, and allows AI systems to track and explain their own decision-making processes in a manner understandable to users and stakeholders. Finally, we discuss the implications of our findings for future research in auditable, human-interpretable AI, as well as the potential ethical considerations of developing AI systems with the appearance of introspective capabilities.

\section{Active inference and introspection}

\subsection{A brief introduction to active inference}

Active inference offers a comprehensive framework for naturalizing, explaining, simulating, and understanding the mechanisms that underwrite decision-making, perception, and action \cite{constant2019regimes, da2021bayesian}. The free energy principle (FEP) is a variational principle of information physics \cite{ramstead2022bayesian}. It has gained considerable attention and traction since it was first introduced in the context of computational neuroscience and biology \cite{friston2010is, friston2005theory}. Active inference denotes a family of models premised on the FEP, which are used to understand and predict the behavior of self-organizing systems. The tools of active inference allow us to model self-organizing systems as driven by the imperative to minimize surprise, which quantifies the degree to which a given path or trajectory deviates from its inertial or characteristic path---or its upper bound, variational free energy, which scores the difference between its predictions and the actual sensory inputs it receives \cite{ramstead2018answering}. 

Active inference modeling work suggests that decision-making, perception, and action involve the optimization of a world model that represents the causal structure of the system generating outcomes of observations \cite{ramstead2022bayesian}. In particular, active inference models the way that latent states or factors in the world cause sensory inputs, and how those factors cause each other, thereby capturing the essential causal structure of the measured or sensed world \cite{konaka2022decoding}. Minimizing surprise or free energy on average and over time allows the brain to maintain a consistent and coherent internal model of the world---one that maximizes predictive accuracy while minimizing model complexity---which, in turn, enables agents to adapt and survive in their environments \cite{friston2010is, friston2013life}. (Strictly speaking, this is the other way around. In other words, agents who ``survive'' can always be read as minimizing variational free energy or maximizing their marginal likelihood (a.k.a., model evidence). This is often called self-evidencing \cite{hohwy2016self}.)

Active inference has instrumental value in allowing us to model, and thereby hopefully help to understand, core aspects of human consciousness (for a review, see \cite{friston2010is}Friston, 2010). Of particular interest to us here, it enables us to model the processes involved in introspective self-access (see \cite{ramstead2023inner, ramstead2023mum}. Active inference modeling deploys the construct of generative models to make sense of the dynamics of self-organizing systems. In this context, a generative model is a joint probability density over the hidden or latent causes of observable outcomes; see \cite{ramstead2022bayesian} for a discussion of how to interpret these models philosophically and \cite{sandved2021towards} for a gentle introduction to the technical implementation of these models.

\begin{figure}[t!]
    \centering
    \includegraphics[width=0.8\textwidth]{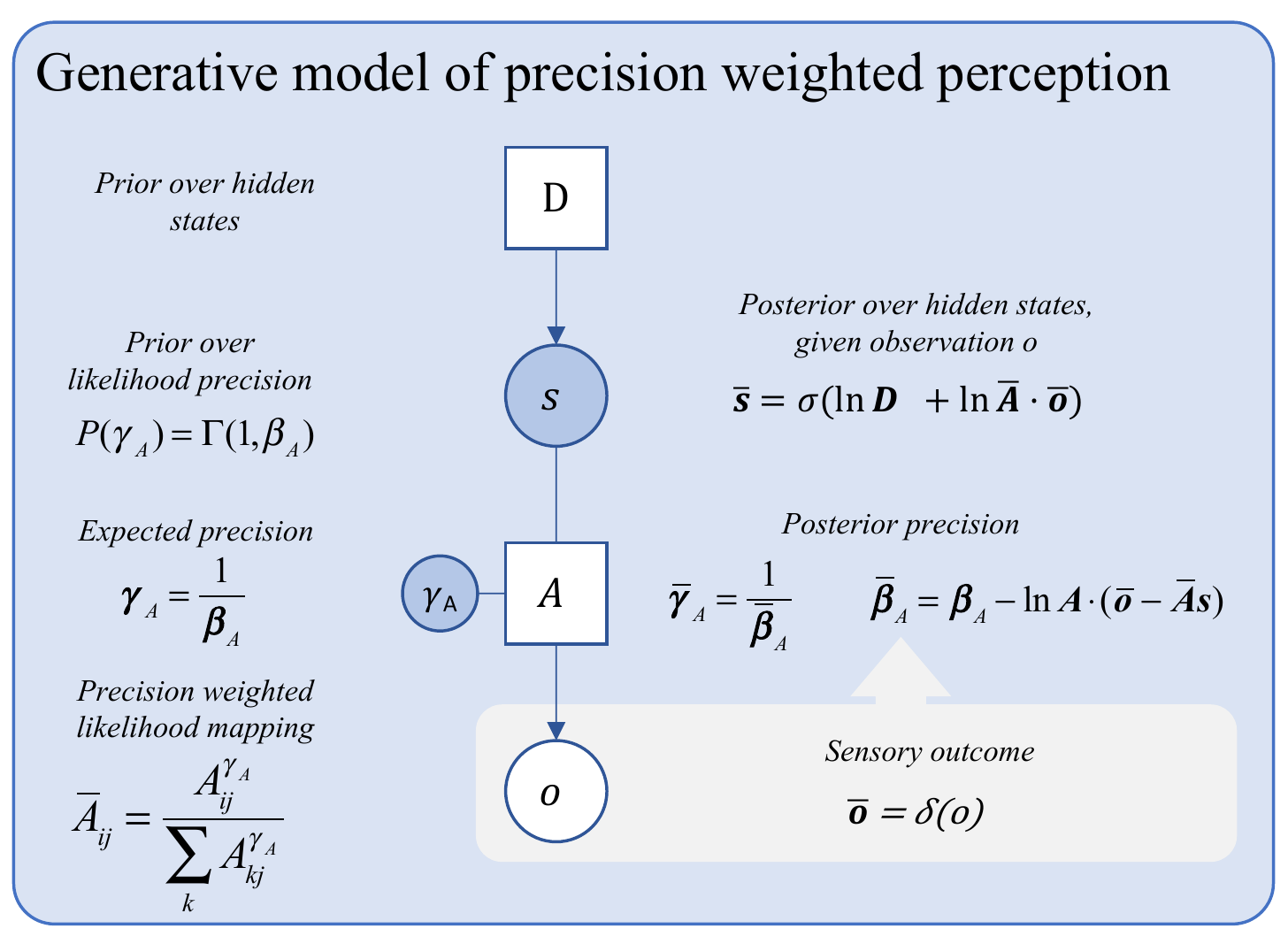}
    \caption{\textbf{A basic generative model for precision-weighted perceptual inference.} This figure depicts an elementary generative model that is capable of performing precision-weighted perceptual inference. States are depicted as circles and denoted in lowercase: observable states or outcomes are denoted $o$ and latent states (which need to be inferred) are denoted $s$. Parameters are depicted as squares and denoted as uppercase. The likelihood mapping \textbf{A} relates outcomes to the states that cause them, whereas \textbf{D} harnesses our prior beliefs about states, independent of how they are sampled. The precision term $\gamma$ controls the precision or weighting assigned to elements of the likelihood, and implements attention as precision-weighting. Figure from \cite{sandved2021towards}.}
    \label{fig:fig1}
\end{figure}

\begin{figure}[t!]
    \centering
    \includegraphics[width=0.8\textwidth]{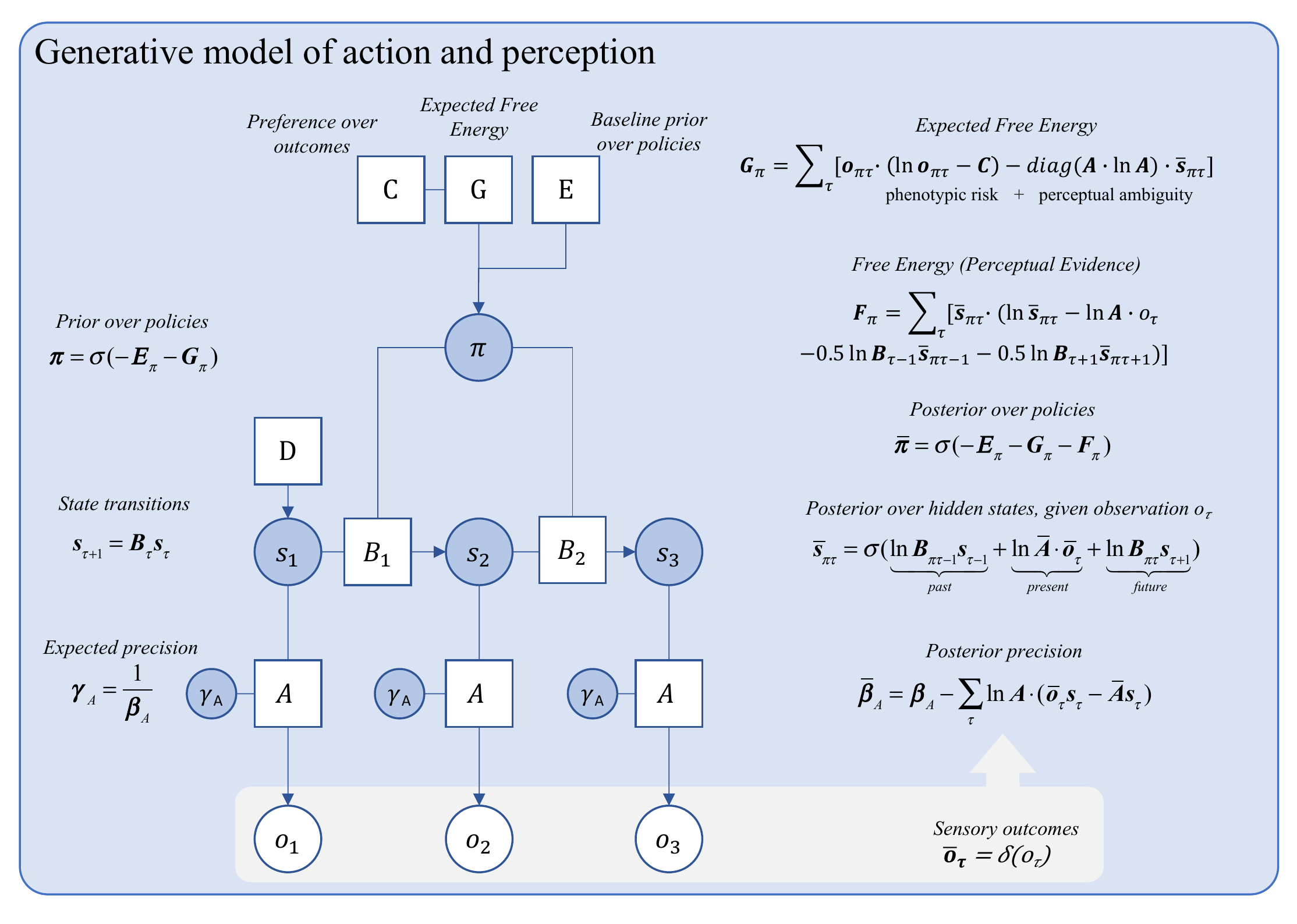}
    \caption{\textbf{A generative model for policy selection.} This figure depicts a more sophisticated generative model that is apt for planning and the selection of actions in the future. The basic model depicted in Figure \ref{fig:fig1} has now been expanded to include beliefs about the current course of action or policy (denoted $\bar{\pi}$), as well as \textbf{B}, \textbf{C}, \textbf{E}, \textbf{F} and \textbf{G} parameters. This kind of model generates a time series of states ($s_1$, $s_2$, etc.) and outcomes ($o_1$, $o_2$, etc.). The state transition (\textbf{B}) parameter encodes the transition probabilities between states over time, independently of the way they are sampled. \textbf{B}, \textbf{C}, \textbf{E}, \textbf{F} and \textbf{G} enter into the selection of beliefs about courses of action, a.k.a.~ policies. The \textbf{C} vector specifies preferred or expected outcomes and enters into the calculation of variational (\textbf{F}) and expected (\textbf{G}) free energies. The \textbf{E} vector specifies a prior preference for specific courses of action. Figure from \cite{sandved2021towards}.}
    \label{fig:fig2}
\end{figure}

We depict a simple generative model, apt for perceptual inference, in Figure \ref{fig:fig1}, and a more complex generative model, apt for the selection of actions (a.k.a.~policy selection) in Figure \ref{fig:fig2}. These models specify the way in which observable outcomes are generated by (typically non-observable) states or factors in the world.

The main advantage of using generative models over current state of the art black box approaches is interpretability and auditability. Indeed, the factors that figure in the generative model are explicitly labeled, such that their contributions to the operations of the model can be read directly off its structure. This lends the generative model a degree of auditability that other approaches do not have.

\subsection{Active inference, introspection, and self-modeling}

Active inference modeling has been deployed in the context of the scientific study of introspection, self-modeling, and self-access, which has led to the development of several leading theories of consciousness (for a review, see \cite{seth2022theories, ramstead2023mum}). Introspection, which is defined as the ability to access and evaluate one's own mental states, thoughts, and experiences, plays a pivotal role in self-awareness, learning, and decision-making and is a pillar of human consciousness \cite{limanowski2018seeing}. Self-modeling and self-access can be defined as interconnected processes that contribute to the development of self-awareness and to the capacity for introspection. Self-modeling involves the creation of internal representations of oneself, while self-access refers to the ability to access and engage with these representations for self-improvement and learning \cite{murray2018self, baker2022going}. These processes, in conjunction with introspection, form a complex dynamic system that enriches our understanding of consciousness and the self—and indeed, may arguably form the causal basis of our capacity to understand ourselves and others.

Introspective self-access has been modeled using active inference by deploying a hierarchically structured generative model \cite{limanowski2020attenuating}. The basic idea is that for a system to report or evaluate its own inferences, it must be able to enact some form of self-access, where some parts of the system can take the output of other parts as their own input, for further processing. This has been discussed in computational neuroscience under the rubric of ``opacity'' and ``transparency'' \cite{metzinger2003phenomenal, metzinger2007empirical, metzinger2017problem, sandved2021towards}. The idea is that some cognitive processes are ``transparent'': like a (clean, transparent) window, they enable us to access some other thing (say, a tree outside) while not themselves being perceivable. Other cognitive processes are ``opaque'': they can be assessed per se, as in introspective self-awareness (i.e., aware that you are looking at a tree as opposed to seeing a tree). The idea, then, is that introspective processes make other cognitive processes accessible to the system as such, rendering them opaque. 

\begin{figure}[t!]
    \centering
    \includegraphics[width=0.8\textwidth]{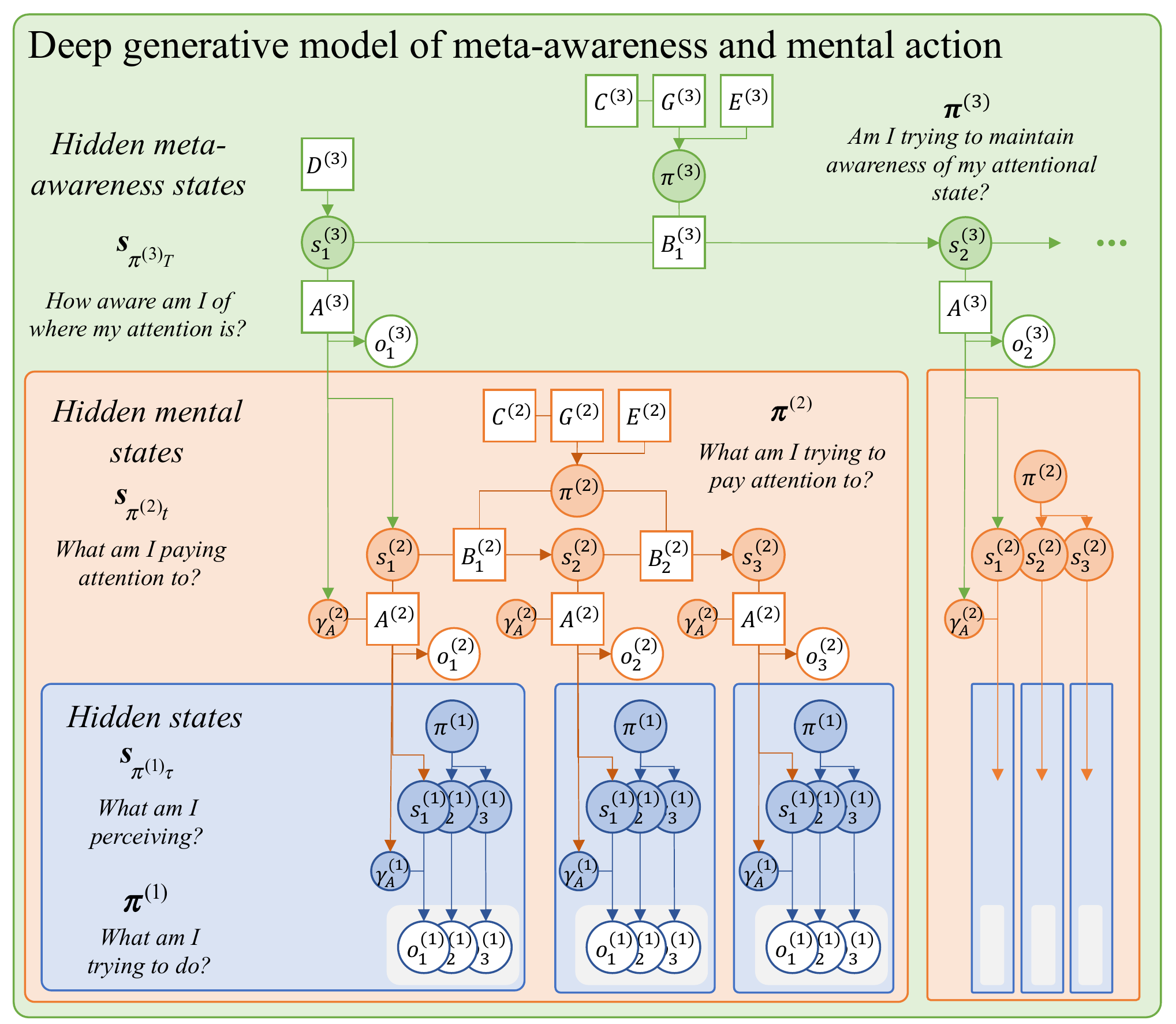}
    \caption{\textbf{A hierarchical generative model capable of self-access.} Here, the generative model depicted in Figure \ref{fig:fig2} (in blue) has been augmented with two superordinate hierarchical layers. In this architecture, posterior state estimates at one level are passed onto the next level as data for further inference. Note that this induces an architecture where the system is able to make inferences about its own inferences. Figure from \cite{sandved2021towards}.}
    \label{fig:fig3}
\end{figure}

In the context of self-access, the transparency and opacity of introspective processes has been modeled using a three-level generative model \cite{sandved2021towards}. The model is depicted in Figure \ref{fig:fig3}. This model provides a framework for understanding how we access and interpret our internal states and experiences. The first level of the model (in blue), which implements the selection of overt actions, can be seen as a transparent process. The second, hierarchically superordinate level (in orange), which implements attention and covert action \cite{metzinger2017problem, ramstead2023inner}, represents more opaque processes, which make processes in the first layer accessible to the system. This layer models mental actions and shifts in attention that we may not be consciously aware of, or able to report. The second level takes as its input the inferences (posterior state estimations) ongoing at the first level, as data for further inference---about the system’s inferences. Attentional processes are of this sort: they are about cognitive processes and action, and they modulate the activity of the first level. The third, final level (in green) implements the awareness of where one’s attention is deployed. In other words, it both recognizes and instantiates a particular attentional set via bottom-up and top-down messages between levels, respectively. On the whole, this three-level architecture models our self-access and introspective abilities in terms of the processes regulating transparency and opacity at a phenomenal level of description, or attentional selection at a psychological level.

Ramstead, Albarracin et al.~(\citeyear{ramstead2023inner}) \cite{ramstead2023inner} recently discussed how active inference enables us to model both overt and covert action (also see \cite{metzinger2017problem, limanowski2018seeing, limanowski2020attenuating, fleming2020awareness, yon2021precision}). Overt actions---observable behaviors such as physical movements or verbal responses---are directly influenced by the brain's hierarchical organization and can be modeled using active inference \cite{friston2011action, friston2017deep, Friston2017a}. In contrast, covert actions refer to internal mental processes, such as attention and imagination, which involve the manipulation and processing of internal representations in the absence of observable behaviors \cite{pezzulo2012active, feldman2010attention, edwards2012bayesian, hohwy2012attention, brown2013active, kanai2015cerebral, vossel2015cortical, ainley2016bodily, limanowski2017dis, parr2019attention}---of the sort discussed as ``mental action'' \cite{metzinger2017problem, limanowski2018seeing, limanowski2022precision, sandved2021towards}. These actions are essential for higher cognitive functions, which rely on the brain’s capacity to explore and manipulate abstract concepts and relationships.

In Smith et al.~(\citeyear{smith2019neurocomputational}) \cite{smith2019neurocomputational}, a hierarchical architecture of this type was deployed that was augmented with the capacity to report on its emotional states. Thus, it is possible to use active inference to design systems that can not only access their own states and perform inferences on their basis, but also to report on their introspective processes in a manner that is readily understandable by human users and stakeholders. With this formulation of how active inference enables agents to model their overt and covert action, in the following sections, we argue that we can and ought to research, design, and develop AI systems that mimic these introspective processes, ultimately leading to more human-like artificial intelligence.

\section{Using active inference to design self-explaining AI}

We argue that incorporating the design principles of active inference into AI systems can lead to better explainability. This is for two key reasons. The first is that, by deploying an explicit generative model, AI systems premised on active inference are designed explicitly such that their operations can be interpreted and audited by a user or stakeholder that is fluent in the operation of such models. We believe that the inherent explainability of active inference AI might be scaled up, by deploying the kind of explicit, standardized world modelling techniques that are being developed as open standards within the Institute of Electrical and Electronics Engineers (IEEE) P2874 Spatial Web Working Group \cite{SpatialWebWorkingGroup}, to formalize contextual relationships between entities and processes and to create digital twins of environments that are able to update in real time.

The second is that, by implementing an architecture inspired by active inference models of introspection, we can build systems that are able to access---and report on---the reasons for their decisions, and their state of mind when reaching these decisions.

AI systems designed using active inference can incorporate the kind of hierarchical self-access described by \cite{sandved2021towards} and by \cite{smith2019neurocomputational}, to enhance their introspection during decision-making. As discussed, in the active inference tradition, introspection can be understood in the context of the (covert and overt) actions that AI systems perform. Covert actions, which are internal computations and decision-making processes that are not directly observable to users and stakeholders, can be recorded or explained to make the system more explainable. Overt actions, which are actions that an AI system takes based on its internal computations, such as making a recommendation or decision, can be explained to help users understand why the AI system acted as it did. This kind of deep inference promotes introspection, adaptability, and responses to environmental changes \cite{dhulipala2022efficient, schoeffer2023interdependence}.

The proposed AI architecture includes components that continuously update and maintain an internal model of its own states, beliefs, and goals. This capacity for self-access (and implicitly self-report) enables the AI system to optimize (and report on) its decision-making processes, fostering introspection (and enhanced explainability). It incorporates metacognitive processing capabilities, which involve the ability to monitor, control, and evaluate its own cognitive processes. The AI system can thereby better explain the factors that contribute to its decisions, as well as identify potential biases or errors, ultimately leading to improved decision-making and explainability.

The proposed AI architecture would include introspection and a self-report interface, which translates the AI system's internal models and decision-making processes into human-understandable (natural) language (using, e.g., large language models). In effect, the agent would be talking to itself, describing its current state of mind and beliefs. This interface bridges the gap between the AI system's internal workings and human users, promoting epistemic trust and collaboration. In this way, the system can effectively mimic human-like consciousness and transparent introspection, leading to a deeper understanding of its decision-making processes and explainability. This advancement may be essential in fostering trust and collaboration between AI systems and their human users, paving the way for more effective and responsible AI applications.

Augmenting a generative model with black box systems---like large language models---may be a useful strategy to help AI systems articulate their ``understanding'' of the world. Using large language models to furnish an introspective interface may be relatively straightforward, leveraging their powerful natural language processing capabilities to create explanations of belief updating. This architecture---with a hierarchical generative model at its core---may contribute to the overall performance and explainability of hybrid AI systems. Attention mechanisms also achieve this purpose by enhancing the explainability of the AI system's decision making, emphasizing important factors in the hierarchical generative model that contribute to its decisions and actions.

These ideas are not new. Attentional mechanisms, particularly those at the word-level, have been identified as crucial components in AI architecture, specifically in the context of hierarchical generative models---and in generative AI, in the form of transformers. They function by focusing on relevant aspects during decision-making processes, thereby allowing the system to effectively process and prioritize information \cite{lan2020kind}. In fact, the performance of hierarchical models, which are a type of AI architecture, can be significantly improved by integrating word-level attention mechanisms. These mechanisms are powerful because they can leverage context information more effectively, especially fine-grained information.

The AI architecture that we propose employs a soft attention mechanism, which uses a weighted combination of hierarchical generative model components to focus on relevant information. The attention weights are dynamically computed based on the input data and the AI system's internal state, allowing the system to adaptively focus on different aspects of the hierarchical generative model \cite{kulkarni2020soft}. This approach is similar to the use of deep learning models for global coordinate transformations that linearize partial differential equations, where the model is trained to learn a transformation from the physical domain to a computational domain where the governing partial differential equations are simpler, or even linear \cite{gin2021deep}.

The AI architecture that we describe here effectively integrates diverse information sources for decision-making, mirroring the complex information processing capabilities observed in the human brain. The hierarchical structure of the generative model facilitates the exchange of information between different levels of abstraction. This exchange allows the AI system to refine and update its internal models based on both high-level abstract knowledge and low-level detailed information. 

In conclusion, the integration of introspective processes in AI systems may represent a significant step towards achieving more explainable AI. By leveraging explicit generative models, as well as attention and introspection mechanisms, we can design AI systems that are not only more efficient and robust, but also more understandable and trustworthy. This approach allows us to bridge the gap between the complex internal computations of AI systems and the human users who interact with them. Ultimately, the goal is to create AI systems that can effectively communicate the reasons that drive their decision-making processes, adapt to environmental changes, and collaborate seamlessly with human users. As we continue to advance in this field, the importance of introspection in AI will only become more apparent, paving the way for more sophisticated and ethically sound AI systems.

\section{Discussion}

\subsection{Directions for future research}

The problem of explainable AI is the problem of understanding how AI models arrive at their decisions or predictions. This problem is especially relevant to avoid biases and harm in the design, implementation, and use of AI systems. By incorporating explicit generative models and introspective processing into the proposed AI architecture, we can create a system that is or seems capable of introspection and, thereby, that displays greatly enhanced explainability and auditability. This approach to AI design paves the way for more effective AI deployment across various real-world applications, by shedding light upon the problem of explainability, thereby offering opportunities for fostering trust, fairness, and inclusivity.

The development of the AI architecture based on active inference opens several potential avenues for future research. One possible direction is to further investigate the role of attention and introspection mechanisms in both AI systems and human cognition, as well as the development of more efficient attentional models to improve the AI system's ability to focus on salient information during decision-making. The approach that we propose bridges the gap between AI and cognitive neuroscience by incorporating biologically-inspired mechanisms into the design of AI systems. As a result, the proposed architecture promotes a deeper understanding of the nature of cognition and its potential applications in artificial intelligence, thus paving the way for more human-like AI systems capable of introspection and enhanced collaboration with human users.

Future work could explore more advanced data fusion techniques, such as deep learning-based fusion or probabilistic fusion, to improve the AI system's ability to combine and process multimodal data effectively. Evaluating the effectiveness of these techniques in diverse application domains will also be a valuable avenue for research \cite{lahat2015multimodal, microsoft2020seeing}. Furthermore, the explanation dimension of these AI systems has been a significant topic in recent years, particularly in decision-making scenarios. These systems provide more awareness of how AI works and its outcomes, building a relationship with the system and fostering trust between AI and humans \cite{ferreira2021human}.

In addition to the aforementioned avenues for future research, another promising direction lies in the realm of computational phenomenology (for a review and discussion, see \cite{ramstead2022generative}. Beckmann, Köstner, \& Hipólito (\citeyear{beckmann2023rejecting}) \cite{beckmann2023rejecting} have proposed a framework that deploys phenomenology---the rigorous descriptive study of first-person experience---for the purposes of machine learning training. This approach conceptualizes the mechanisms of artificial neural networks in terms of their capacity to capture the statistical structure of some kinds of lived experience, offering a unique perspective on deep learning, consciousness, and their relation. By grounding AI training in socioculturally situated experience, we can create systems that are more aware of sociocultural biases and capable of mitigating their impact. Ramstead et al.~(\citeyear{ramstead2022generative}) \cite{ramstead2022generative} propose a similar methodology based on explicit generative models as they figure in the active inference tradition. This connection to first-person experience, of course, does not guarantee unbiased AI. But by moving away from traditional black box AI systems, we shift towards human-interpretable models that enable the identification and correction of biases in the AI system. This approach aligns with our goal of creating AI systems that are not only efficient and effective, but also ethically sound and socially responsible.

The incorporation of computational phenomenology into our proposed AI architecture could further enhance its introspective capabilities and its ability to understand and navigate the complexities of human sociocultural contexts. This could lead to AI systems that are more adaptable, more trustworthy, and more capable of meaningful collaboration with human users. As we continue to explore and integrate such innovative approaches, we move closer to our goal of creating AI systems that truly mirror the richness and complexity of human cognition and consciousness.

\subsection{Ethical considerations of introspective AI systems}

Ethical AI starts with the development of AI systems that are ethically designed; AI systems must be designed in such a way as to be transparent, auditable, and explainable, and to minimize harm. But as these systems become increasingly integrated into our daily lives, research on the ethical implications of introspective AI systems, as well as the development of regulatory frameworks and guidelines for responsible AI use, become crucial. The development of introspective AI systems raises several ethical considerations. Even if these systems provide more human-like decision-making capabilities and enhanced explainability, it is and will remain crucial to ensure that their decisions are transparent, fair, and unbiased, and that their designers and users can be held accountable for harm that their use may cause.

To address these concerns, future research should focus on developing methods to audit and evaluate the AI system's decision-making processes, as well as identify and mitigate potential biases within the system. Additionally, the development of ethical guidelines and regulatory frameworks for the use of introspective AI systems will be essential to ensure that they are deployed responsibly and transparently. Moreover, as introspective AI systems become more prevalent, issues related to agency, privacy, and data security may arise. Ensuring that these systems protect sensitive information by abiding by data protection regulations, thereby safeguarding agency, will be of paramount importance.

In conclusion, the development of AI systems based on active inference has broad implications for both the fields of AI and consciousness studies. As future research explores the potential of this novel approach, ethical considerations and responsible use of introspective AI systems must remain at the forefront of these advancements, ultimately leading to more transparent, effective, and user-friendly AI applications.

\section{Conclusion}

We have argued that active inference has demonstrated significant potential in advancing the field of explainable AI. By incorporating design principles from active inference, the AI system can better tackle complex real-world problems with improved auditability of decision-making, thereby increasing safety and user trust.

Throughout our discussions and analysis, we have highlighted the importance of active inference models as a foundation for designing more human-like AI systems, seemingly capable of introspection and finessed (epistemic) collaboration with human users. This novel approach bridges the gap between AI and cognitive neuroscience by incorporating biologically-inspired mechanisms into the design of AI systems, thus promoting a deeper understanding of the nature of consciousness and its potential applications in artificial intelligence.

As we move forward in the development of AI systems, the importance of advancing explainable AI becomes increasingly apparent. By designing AI systems that can not only make accurate and efficient decisions, but also provide understandable explanations for their decisions, we foster (epistemic) trust and collaboration between AI systems and human users. This advancement ultimately leads to more transparent, effective, and user-friendly AI applications that can be tailored to a wide range of real-world scenarios.

\begin{sloppypar}
\printbibliography[title={References}]
\end{sloppypar}

\end{document}